\documentclass[runningheads]{llncs}
\usepackage{graphicx}
\usepackage{amsmath,amssymb} 
\usepackage{color}

\usepackage{multirow}
\usepackage{lipsum}

\DeclareMathOperator*{\argmaxB}{argmax   }   
\newcommand\blfootnote[1]{%
  \begingroup
  \renewcommand\thefootnote{}\footnote{#1}%
  \addtocounter{footnote}{-1}%
  \endgroup
}
\begin{document}
\title{Ask, Acquire, and Attack: Data-free UAP Generation using Class Impressions} 

\titlerunning{Ask, Acquire, and Attack}
%
\author{Konda Reddy Mopuri* \and
Phani Krishna Uppala* \and
R. Venkatesh Babu}
%
\authorrunning{K.R. Mopuri, P.K. Uppala and R.V. Babu}
%

\institute{Video Analytics Lab, Indian Institute of Science, Bangalore, India \\
\email{kondamopuri@iisc.ac.in, krishnaphaniiitg@gmail.com, venky@iisc.ac.in}
}
\maketitle              
\begin{abstract}
\blfootnote{*Equal contribution}Deep learning models are susceptible to input specific noise, called adversarial perturbations. Moreover, there exist input-agnostic noise, called Universal Adversarial Perturbations (UAP) that can affect inference of the models over most input samples. Given a model, there exist broadly two approaches to craft UAPs: (i) data-driven: that require data, and (ii) data-free: that do not require data samples. Data-driven approaches require actual samples from the underlying data distribution and craft UAPs with high success (fooling) rate. However, data-free approaches craft UAPs without utilizing any data samples and therefore result in lesser success rates. In this paper, for data-free scenarios, we propose a novel approach that emulates the effect of data samples with class impressions in order to craft UAPs using data-driven objectives. Class impression for a given pair of category and model is a generic representation (in the input space) of the samples belonging to that category. Further, we present a neural network based generative model that utilizes the acquired class impressions to learn crafting UAPs. Experimental evaluation demonstrates that the learned generative model, (i) readily crafts UAPs via simple feed-forwarding through neural network layers, and (ii) achieves state-of-the-art success rates for data-free scenario and closer to that for data-driven setting without actually utilizing any data samples.

\end{abstract}
\section{Introduction}
\label{sec:introduction}
Machine learning models are pregnable (e.g. \cite{prsystemsunderattack-pari-2014,evasion-mlkd-2013,adversarialml-acmmm-2011}) at test time to specially learned, mild noise in the input space, commonly known as adversarial perturbations. Data samples created via adding these perturbations to clean samples are known as adversarial samples. Lately, the Deep Neural Networks (DNN) based object classifiers are also observed~\cite{intriguing-iclr-2014,explainingharnessing-iclr-2015,deepfool-cvpr-2016,physicalworld-iclr-2017} to be drastically affected by the adversarial attacks with quasi imperceptible perturbations. Further, it is observed (e.g.~\cite{intriguing-iclr-2014}) that these adversarial perturbations exhibit cross model generalizability (transferability). This means, often same adversarial sample gets incorrectly classified by multiple models in spite of having different architectures and trained with disjoint training datasets. It enables attackers to launch simple black-box attacks~\cite{practicalbb-asiaccs-2017,delving-iclr-2017} on the deployed models without any knowledge about their architecture and parameters.

However, most of the existing works (e.g.~\cite{intriguing-iclr-2014,deepfool-cvpr-2016}) craft input-specific perturbations, i.e., perturbations are functions of input and they may not transfer across data samples. In other words, perturbation crafted for one data sample most often fails to fool the model when used to corrupt other clean data samples. However, recent findings by Moosavi-Dezfooli \textit{et al.}~\cite{universal-cvpr-2017} and Mopuri \textit{et al.}~\cite{mopuri-bmvc-2017,mopuri-pami-2018} demonstrated that there exist input-agnostic (or image-agnostic) perturbations that when added, most of the data samples can fool the target classifier. Such perturbations are known as ``Universal Adversarial Perturbations (UAP)", since a single noise can adversarially perturb samples from multiple categories. Furthermore, it is observed that similar to image-specific perturbations, UAPs also exhibit cross model generalizability enabling easy black-box attacks. Thus,
UAPs pose a severe threat to the deployment of the vision models and require a meticulous study. Especially for applications which involve safety (e.g. autonomous driving) and privacy of the users (e.g. access granting), it is indispensable to develop robust models against such adversarial attacks.

\begin{figure}[t]
        \centering
        \includegraphics[width=\textwidth]{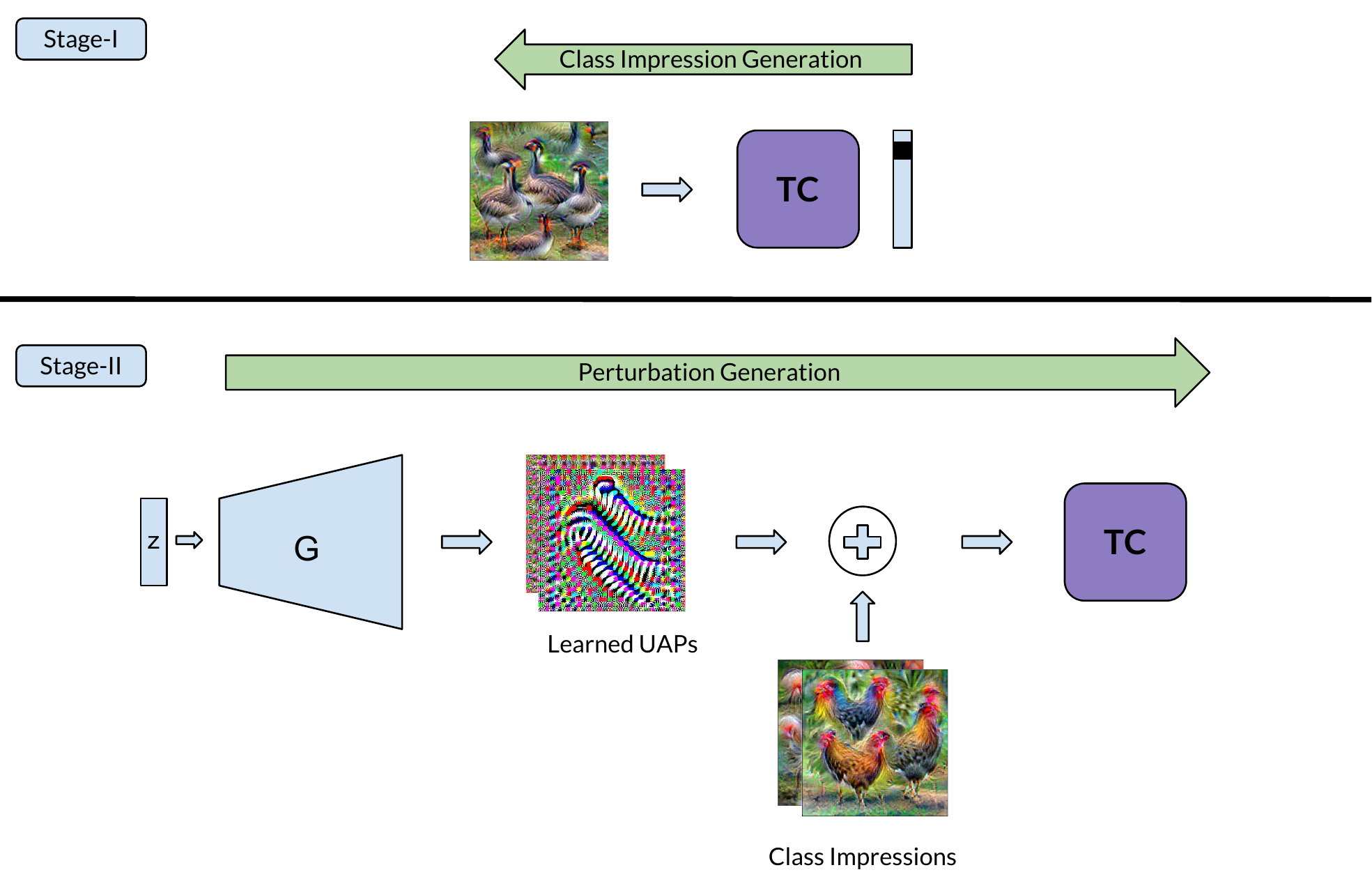}
        \caption{Overview of the proposed approach. Stage-I, ``Ask and Acquire" generates the ``class impressions" to mimic the effect of actual data samples. Stage-II, ``Attack" learns a neural network based generative model $G$ which crafts UAPs from random vectors $z$ sampled from a latent space.
        }
        \label{fig:overview}        
\end{figure}

Approaches that craft UAPs can be broadly categorized into two classes: (i) data-driven, and (ii) data-free approaches. Data-driven approaches such as~\cite{universal-cvpr-2017} require access to samples of the underlying data distribution to craft UAPs using a fooling objective (e.g. confidence reduction as in eq~(\ref{eqn:l_f})). Thus, UAPs crafted via data-driven approaches typically result in higher success rate (or fooling rate), i.e., fool the models more often. Note that data-driven approaches have access to the data samples and the model architecture along with the parameters. Further, performance of the crafted UAPs is observed (\cite{mopuri-bmvc-2017,mopuri-pami-2018}) to be proportional to the number of data samples available during crafting. However the data-free approaches (e.g. FFF~\cite{mopuri-bmvc-2017}), with a goal to understand the true stability of the the models, indirectly craft UAPs (e.g. activation loss of FFF~\cite{mopuri-bmvc-2017}) instead of using a direct fooling objective. Note that data-free approaches have access to only the model architecture and parameters but not to any data samples. Thus, it is a challenging problem to craft UAPs in data-free scenarios and therefore the success rate of these UAPs would typically be lesser compared to that achieved by the data-driven ones.

In spite of being difficult, data-free approaches have important advantages:
\begin{itemize}
    \item When compared to their data-driven counter parts, data-free approaches reveal accurate vulnerability of the learned representations and in turn the models. On the other hand, success rates reported by data-driven approaches act as a sort of upper bounds on the achievable rates. Also, it is observed (\cite{mopuri-bmvc-2017,mopuri-pami-2018}) that their performance is proportional to the amount of data available for crafting UAPs.
    \item Because of the strong association of the data-driven UAPs to the target data, they suffer poor transferability across datasets. On the other hand, data-free UAPs transfer better across datasets~\cite{mopuri-bmvc-2017,mopuri-pami-2018}.
    \item Data-free approaches are typically faster~\cite{mopuri-bmvc-2017} to craft UAPs.
\end{itemize}

Thus, in this paper, we attempt to achieve best of both worlds, i.e.,
effectiveness of the data-driven objectives and efficiency, transferability of the data-free approaches. We present a novel approach for the data-free scenarios that emulates the effect of actual data samples with ``\textit{class impressions}" of the model and crafts UAPs via learning a feed-forward neural network. Class impressions are the reconstructed images from the model's memory which is the set of learned parameters. In other words, they are generic representations of the object categories in the input space (as shown in Fig.~\ref{fig:class_impressions}). In the first part of our approach, we acquire class impressions via simple optimization (sec.~\ref{subsec:class_impressions}) that can serve as representative samples from the underlying data distribution. After acquiring multiple class impressions for each of the categories, we perform the second part, which is learning a generative model (a feed-forward neural network) for efficiently generating UAPs. Thus, unlike the existing works (\cite{universal-cvpr-2017,mopuri-bmvc-2017}) that solve complex optimizations to generate UAPs, our approach crafts via a simple feed-forward operation through the learned neural network. The major contributions of our work can be listed as:
\begin{itemize}
    \item We propose a novel approach to handle the absence of data (via class impressions, sec.~\ref{subsec:class_impressions}) for crafting UAPs and achieve state-of-the-art success (fooling) rates.
    \item We present a generative network (sec.~\ref{subsec:crafting}) that learns to efficiently generate UAPs utilizing the class impressions.
\end{itemize}
The paper is organized as followed: section~\ref{sec:related_works} describes the relevant existing works, section~\ref{sec:proposed_approach} presents the proposed framework in detail, section~\ref{sec:experiments} reports comprehensive experimental evaluation of our approach and finally section~\ref{sec:conclusions} concludes the paper.

\section{Related Works}
\label{sec:related_works}
Adversarial perturbations~(e.g. \cite{intriguing-iclr-2014,explainingharnessing-iclr-2015,deepfool-cvpr-2016}) reveal the vulnerability of the learning models to specific noise. Further, these perturbations can be input agnostic~\cite{universal-cvpr-2017,mopuri-bmvc-2017} called ``Universal Adversarial Perturbations (UAP)" and can pose severe threat to the deployability of these models. Existing approaches to craft the UAPs (\cite{universal-cvpr-2017,mopuri-bmvc-2017,mopuri-pami-2018}) perform complex optimizations every time we wish to craft a UAP. Differing from the previous works, we present a neural network that readily crafts UAPs. Only similar work by Baluja~\textit{et al.}~\cite{atn-aaai-2018} presents a neural network that transforms a clean image into an adversarial sample by passing through a series of layers. However, we learn a generative model which maps a latent space to that of UAPs. A concurrent work by Mopuri \textit{et al.}~\cite{mopuri-cvpr-2018} presents a similar generative model approach to craft perturbations but for data-driven case.

Also, existing data-free method~\cite{mopuri-bmvc-2017} to craft UAPs achieves significantly less success rates compared to the data-driven methods such as UAP~\cite{universal-cvpr-2017} and NAG~\cite{mopuri-cvpr-2018}. In this paper, we attempted to reduce the gap between them by emulating the effect of data with the proposed class impressions. Our class impressions are obtained via simple optimization similar to visualization works such as~\cite{backprop-iclrw-2014,guidedbackprop-iclrw-2015}. Feature visualizations~\cite{backprop-iclrw-2014,guidedbackprop-iclrw-2015,deconv-eccv-2014,cam-cvpr-2016,gradcam-iccv-2017,exbp-eccv-2016,mopuri-cnnf-2017} are introduced (i) to understand what input patterns each neuron responds to, and (ii) gain intuitions into neural networks in order to alleviate the black-box nature of the neural networks. Two slightly different approaches exist for feature visualizations. In the first approach, a random input is optimized in order to maximize the activation of a chosen neuron (or set of neurons) in the architecture. This enables to generate visializations for a given neuron (as in~\cite{backprop-iclrw-2014}) in the input space.

In other approaches such as the Deep Dream~\cite{deep-dream-2015} instead of choosing a neuron to activate, arbitrary natural image is passed as an input, and the network enhances the activations that are detected. This way of visualization finds the subtle patterns in the input and amplify them. Since our task is to generate class impressions that emulate the behaviour of real samples, we follow the former approach. 

Since the objective is to generate class impressions that can be used to craft UAPs with the fooling objective, softmax probability neuron seems like the obvious choice to activate. However, this intuition is misleading, \cite{backprop-iclrw-2014,olah2017feature} have shown that directly optimizing at softmax leads to increase in the class probability by reducing the pre-softmax logits of other classes. Also, often it does not increase the pre-softmax value of the desired class, thus giving poor visualizations. In order to make the desired class more likely, we optimize the pre-softmax logits and our observations are in agreement with that of~\cite{backprop-iclrw-2014,olah2017feature}.
\section{Proposed Approach}
\label{sec:proposed_approach}
In this section we present the proposed approach to craft efficient UAPs for data-free scenarios. It is understood (\cite{universal-cvpr-2017,mopuri-bmvc-2017,mopuri-cvpr-2018}) that, because of data availability and a more direct optimization, data-driven approaches can craft UAPs that are effective in fooling. On the other hand, the data-free approaches can quickly craft generalizable UAPs by solving relatively simple and indirect optimizations. In this paper we aim to achieve the effectiveness of the data-driven approaches in the data-free setup. For this, first we create representative data samples called, \textit{class impressions} (Figure~\ref{fig:class_impressions}) to mimic the actual data samples of the underlying distribution. Later, we learn a neural network based generative model to craft UAPs using the generated class impressions and a direct fooling objective (eq.(\ref{eqn:l_f})).  Figure~\ref{fig:overview} shows the overview of our approach. Stage-I, ``Ask and Acquire" is about the class impression generation from the target CNN model and Stage-II, ``Attack" is training the generative model that learns to craft UAPs using the class impressions obtained in the first stage. In the following subsections, we will discuss these two stages in detail. 
\subsection{Notation}
\label{subsec:notation}
We first define the notations followed throughout this paper:
\begin{itemize}
    \item $f$: target classifier (TC) under attack, which is a trained model with frozen parameters
    \item $f^i_k$: $k^{th}$ activation in $i^{th}$ layer of the target classifier
    \item $f^{ps/m}$: output of the pre-softmax layer
    \item $f^{s/m}$: output of the softmax (probability) layer
    \item $v$: additive universal adversarial perturbation (UAP)
    \item $x$: clean input to the target classifier, typically either data sample or class impression
    \item $\xi$: max-norm $(l_1)$ constraint on the UAPs, i.e., maximum allowed strength of perturbation that can be added or subtracted at each pixel in the image
\end{itemize}
\subsection{Ask and Acquire the Class Impressions}
\label{subsec:class_impressions}
Availability of the actual data samples can enable to solve for a direct fooling objective thus craft UAPs that can achieve high success rates~\cite{universal-cvpr-2017}. Hence in the data-free scenarios we generate samples that act as proxy for data. Note that the attacker has access to only the model architecture and the learned parameters of the target classifier (CNN). The learned parameters are a function of training data and procedure. They can be treated as model's memory in which the essence of training has been encoded and saved. The objective of our first stage, ``Ask and Acquire" is to tap the model's memory and acquire representative samples of the training data. We can then use only these representative samples to craft UAPs to fool the target classifier.

Note that we do not aim to generate natural looking data samples. Instead, our approach creates samples for which the target classifier predicts strong confidence. That is, we create samples such that the target classifier strongly believes them to be actual samples that belong to categories in the underlying data distribution. In other words, these are impressions of the actual training data that we try to reconstruct from model's memory. Therefore we name them \textit{Class Impressions}. The motivation to generate these class impression is that, for the purpose of optimizing a fooling objective (e.g. eq.~\ref{eqn:l_f}) it is sufficient to have samples that behave like natural data samples, which is, to be predicted with high confidence. Thus, the ability of the learned UAPs to act as adversarial noise to these samples with respect to the target classifier generalizes to the actual samples. 

Top panel of Fig.~\ref{fig:overview} shows the first stage of our approach to generate the class impressions. We begin with a random noisy image sampled from $\mathcal{U}[0,255]$ and update it till the target classifier predicts a chosen category with high confidence. We achieve this via performing the optimization shown in eq~(\ref{eqn:class-impressions}). Note that we can create impression $(CI_c)$ for any chosen class $(c)$ by maximizing the predicted confidence to that class. In other words, we modify the random (noisy) image till the target network believes it to be an input from a chosen class $c$ with high confidence. We consider the activations in the pre-softmax layer $f_c^{ps/m}$ (before we apply the softmax non-linearity) and maximize the model's confidence.
\begin{equation}
 CI_c = \argmaxB_x \:\:\: f_c^{ps/m}(x) 
 \label{eqn:class-impressions}
\end{equation}
\begin{figure}[t]
\centering
\noindent\begin{minipage}{\textwidth}
  \centering
  \begin{minipage}{.19\textwidth}
  	\centering
    \includegraphics[width=\linewidth]{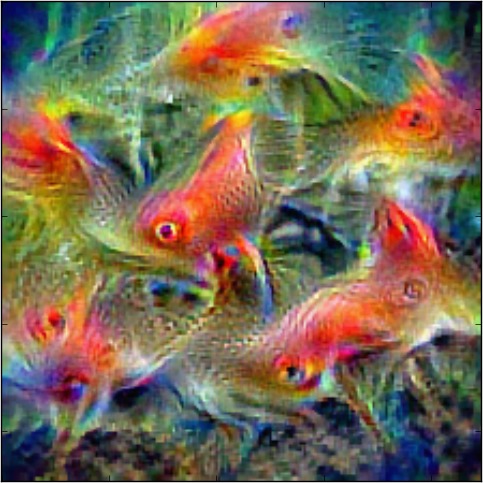}\
    Goldfish
  \end{minipage}
   \begin{minipage}{.19\textwidth}
   	\centering
    \includegraphics[width=\linewidth]{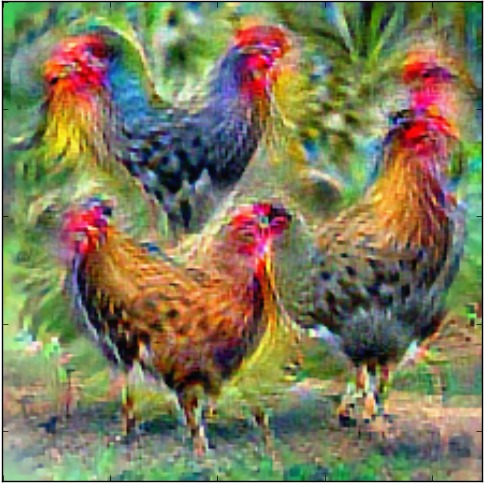}\
    Cock
  \end{minipage}
  \begin{minipage}{.19\textwidth}
  	\centering
    \includegraphics[width=\linewidth]{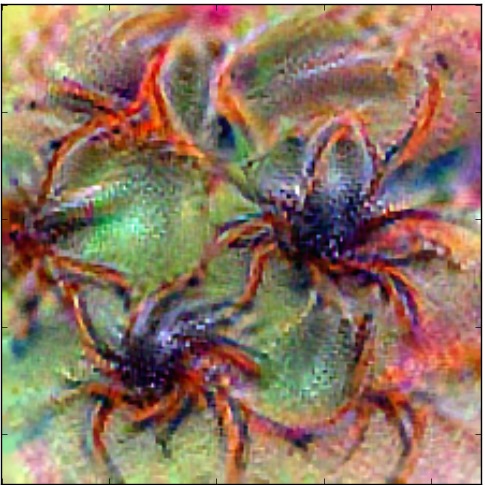}\
    Wolf spider
  \end{minipage}
  \begin{minipage}{.19\textwidth}
  	\centering
    \includegraphics[width=\linewidth]{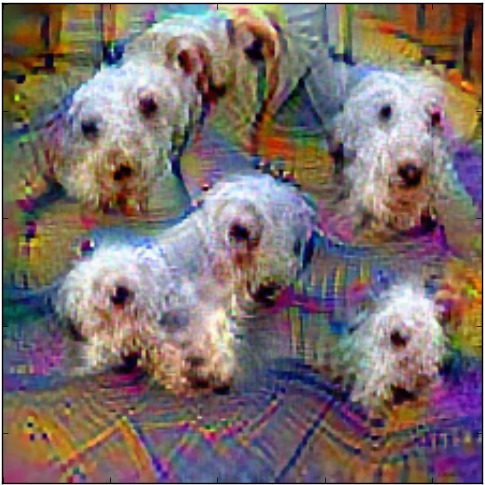}\
    Lakeland terrier
  \end{minipage}
  \begin{minipage}{.19\textwidth}
  	\centering
    \includegraphics[width=\linewidth]{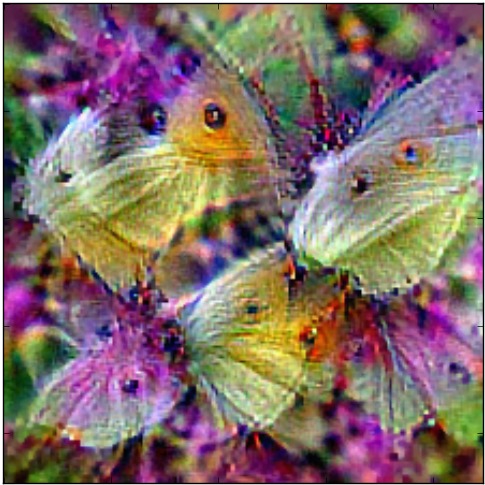}\
    Monarch
  \end{minipage}
\end{minipage}
\caption{Sample class impressions generated for VGG-F~\cite{vggf-bmvc-2014} model. The name of the corresponding categories are mentioned below the images. Note that the impressions have several natural looking patterns located in various spatial locations and in multiple orientations.}
\label{fig:class_impressions}
\end{figure}
While learning the class impressions, we perform typical data augmentations such as (i) random rotation in $[-5^o, 5^o]$, (ii) scaling by a factor randomly selected from $\{0.95, 0.975, 1.0, 1.025\}$, (iii) RGB jittering, and (iv) random cropping. Along with the above typical augmentations, we also add random uniform noise in $\mathcal{U}[-10,10]$. Purpose of this augmentation is to generate robust impressions that behave similar to natural samples with respect to the augmentations and random noise. We can generate multiple impressions for a single category by varying the initialization, i.e., multiple initializations result in multiple class impressions. Note that the dimensions of the generated impressions would be same as that required by the model's input (e.g., $224 \times 224 \times 3$). We have implemented the optimization given in eq~(\ref{eqn:class-impressions}) in TensorFlow~\cite{tensorflow2015-whitepaper-short} framework. We used Adam~\cite{kingma2014adam} optimizer with a learning rate of $0.1$ with other parameters set to their default values. In order to mimic the variety in terms of the difficulty of recognition (from easy to difficult samples), we have devised a stopping criterion for the optimization. We presume that the difficulty is inversely related to the confidence predicted by the classifier. Before we start the optimization in eq.~(\ref{eqn:class-impressions}), we randomly sample a confidence value uniformly in $[0.55,0.99]$ range and stop our optimization after the predicted confidence by the target classifier reaches that. Thus, the generated class impressions will have samples of varied difficulty. 

\begin{figure}[]
\centering
\noindent\begin{minipage}{\textwidth}
  \centering
  \begin{minipage}{.19\textwidth}
  	\centering
    \includegraphics[width=\linewidth]{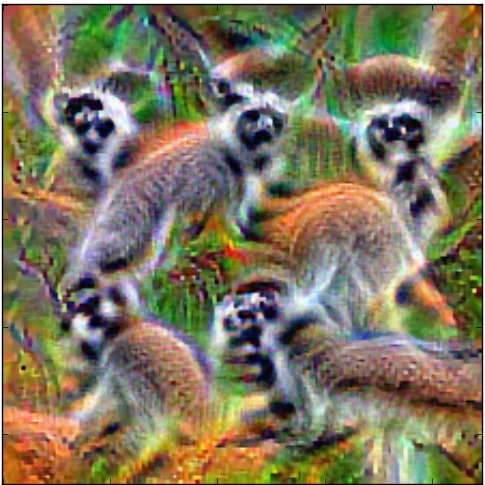}
  \end{minipage}
   \begin{minipage}{.19\textwidth}
   	\centering
    \includegraphics[width=\linewidth]{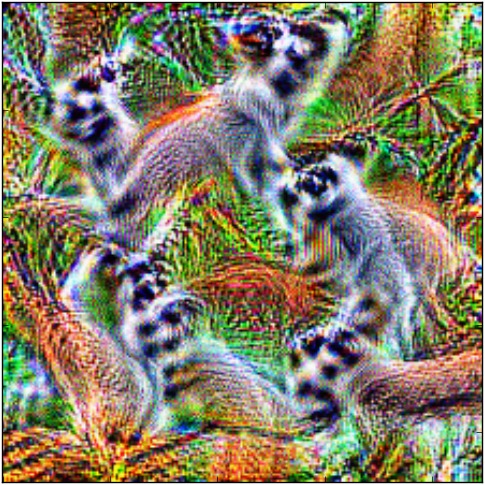}
  \end{minipage}
  \begin{minipage}{.19\textwidth}
  	\centering
    \includegraphics[width=\linewidth]{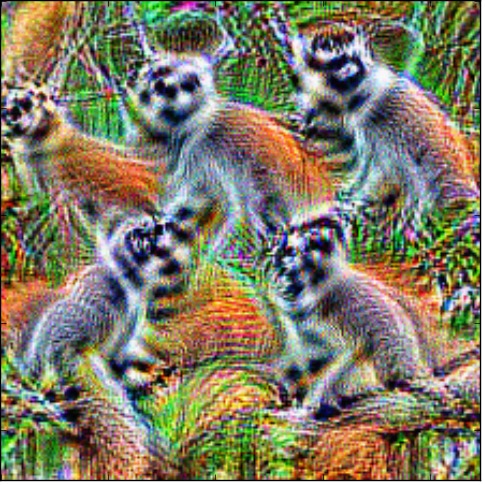}
  \end{minipage}
  \begin{minipage}{.19\textwidth}
  	\centering
    \includegraphics[width=\linewidth]{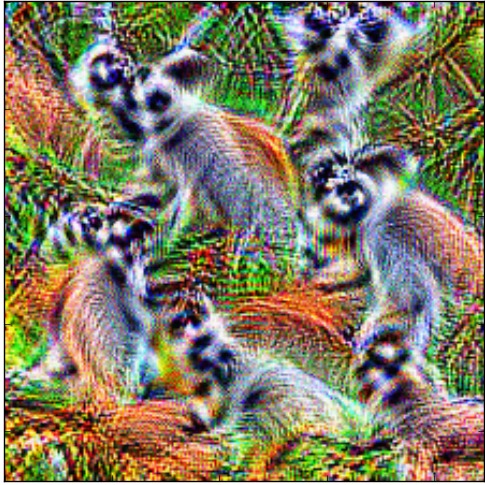}
  \end{minipage}
  \begin{minipage}{.19\textwidth}
  	\centering
    \includegraphics[width=\linewidth]{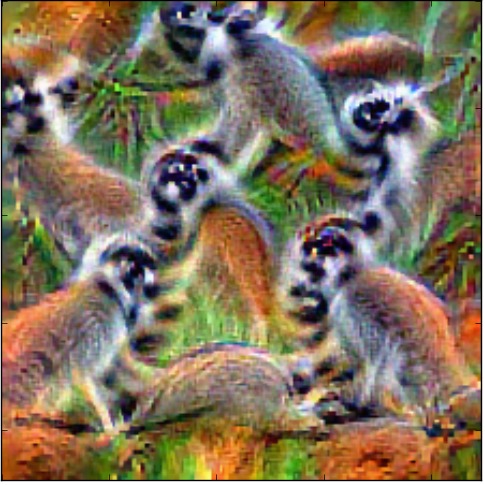}
  \end{minipage}
\end{minipage}
\caption{Multiple class impressions for ``Squirrel Monkey" category generated from different initializations for VGG-F~\cite{vggf-bmvc-2014} target classifier. }
\label{fig:variety_in_class_impressions}
\end{figure}

Fig.~\ref{fig:class_impressions} shows sample class impressions generated for VGG-F~\cite{vggf-bmvc-2014} model. The corresponding category labels are mentioned below the impressions. Note that the generated class impressions clearly show several natural looking patterns located in various spatial locations and in multiple orientations. Fig.~\ref{fig:variety_in_class_impressions} shows multiple class impressions generated by our method starting from different initializations for ``Squirrel Monkey" category. Note that the impressions have different visual patterns relevant to the chosen category. We have generated $10$ class impressions for each of the $1000$ categories in ILSVRC dataset resulting in a total of $10000$ class impressions. These samples will be used to learn a neural network based generative model that can craft UAPs through a feed-forward operation.

\subsection{Attack: Craft the data-free perturbations}
\label{subsec:crafting}
After generating the class impressions in the first stage of our approach, we treat them as training data for learning a generator to craft the UAPs. Bottom panel of Fig.~\ref{fig:overview} shows the overview of our generative model. In the following subsections we present the architecture of our model along with the objectives that drive the learning.
\subsection{Fooling loss}
\label{subsec:fooling-loss}
We learn a neural network $(G)$ similar to the generator part of a Generative Adversarial network (GAN)~\cite{gan-nips-2014}. $G$ takes a random vector $z$ whose components are sampled from a simple distribution (e.g. $\mathcal{U}[-1,1]$) and transforms it into a UAP via a series of deconvolution layers. Note that in practice a mini-batch of vectors is processed. We train $G$ in order to be able to generate the UAPs that can fool the target classifier over the underlying data distribution. To be specific, we train with a fooling loss computed over the generated class impressions (from Stage-I, sec.~\ref{subsec:class_impressions}) as the training data. Let us denote the predicted label on clean sample $(x)$ as `clean label' and that of a perturbed sample $(x+v)$ as `perturbed label'. The objective is to make the `clean' and `perturbed' labels different. To ensure this to happen, our training loss reduces the confidence predicted to the `clean label' on the perturbed sample. Because of the softmax nonlinearity, confidence predicted to some other label increases and eventually causes a label flip, which is fooling the target classifier. Hence, we formulate our fooling loss as

\begin{equation}
    L_f = -log(1-f^{s/m}_c(x+v))
\label{eqn:l_f}
\end{equation}

where $c$ is the clean label predicted on $x$ and $f^{s/m}_c$ is the probability (soft-max output) predicted to category $c$. Note that this objective is similar to most of the adversarial attacking methods (e.g. FGSM~\cite{explainingharnessing-iclr-2015,practicalbb-asiaccs-2017}) in spirit. 
\subsection{Diversity loss}
\label{subsec:diversity_loss}
Fooling loss $L_f$ (eq.(\ref{eqn:l_f})) only trains $G$ to learn UAPs that can fool the target classifier. In order to avoid learning a degenerate $G$ which can only generate a single strong UAP, we enforce diversity in the generated UAPs. We enforce that the crafted UAPs within a mini-batch are diverse via maximizing the pairwise distance between their embeddings $f^l(x+v_i) \text{ and } f^l(x+v_j)$, where $v_i$ and $v_j$ belong to generations within a mini-batch. We consider the layers of the target CNN for projecting $(x+v)$. Thus our training objective is comprised of a diversity loss given by

\begin{equation}
    L_d = -\sum_{i,j=1, i\neq j }^K  d( f^l(x+v_i) , f^l(x+v_j) )
\label{eqn:l_d}
\end{equation}

where $K$ is the mini-batch size, and $d$ is a suitable distance metric (e.g., Euclidean or cosine distance) computed between the features extracted between a pair of adversarial samples. Note that the class impression $x$ present in the two embeddings $f(x+v_i) \text{and} f(x+v_j)$ is same. Therefore, pushing them apart via minimizing $L_d$ will make the UAPs $v_i$ and $v_j$ dissimilar.

Therefore the loss we optimize for training our generative model for crafting UAPs is given by
\begin{equation}
Loss = L_f + \lambda L_d
\label{eqn:loss}
\end{equation}
Note that this objective is similar in spirit to that presented in the concurrent work~\cite{mopuri-cvpr-2018}.
\section{Experiments}
\label{sec:experiments}
In this section we present our experimental setup and the effectiveness of the proposed method in terms the success rates achieved by the crafted UAPs. For all our experiments we have considered ILSVRC~\cite{imagenet-ijcv-2015} dataset and recognition models trained on it as the target CNNs. Note that, since we have considered data-free scenario, we extract class impressions to serve as data samples. Similar to the existing data-driven approach (\cite{universal-cvpr-2017}) that uses $10$ data samples per class, we also extract $10$ impressions for each class which makes a training data of $10000$ samples.
\subsection{Implementation details}
\label{subsec:implementation-details}
The dimension of the latent space is chosen as $10$, i.e, $z$ is random $10D$ vector sampled from $\mathcal{U}[-1,1]$. We have investigated with other dimensions (e.g. $50$, $100$, etc.) for the latent space and found that $10$ is efficient with respect to the number of parameters though the success rates are not very different. We used a mini-batch size of $32$. All our experiments are implemented in TensorFlow~\cite{tensorflow2015-whitepaper-short} using Adam optimizer. The generator part $(G)$ of the network maps the latent space $Z$ to the UAPs for a given target classifier. The architecture of our generator consists of $5$ deconv layers. The final deconv layer is followed by a $tanh$ non-linearity and scaling by $\xi$. Doing so limits the perturbations to $\bigl[-\xi,\: \xi\bigr]$. Similar to~\cite{universal-cvpr-2017,mopuri-bmvc-2017}, the value of $\xi$ is chosen to be $10$ in order to add negligible adversarial noise. The architecture of $G$ is adapted from~\cite{NIPS2016_6125}. We experimented on a variety of CNN architectures trained to perform object recognition on the ILSVRC~\cite{imagenet-ijcv-2015} dataset. The generator $(G)$ architecture is unchanged for different target CNN architectures and separately learned with the corresponding class impressions.

While computing the diversity loss (eq.~\ref{eqn:l_d}), for each of the class impressions in the mini-batch $(x)$, we select a pair of generated UAPs $(v_1 \text{ and } v_2)$ and compute the distance between $f^l(x+v_1) \text{ and } f^l(x+v_2)$. The diversity loss would be sum of all such distances computed over the mini-batch members. We typically consider the softmax layer of the target CNN for extracting the embeddings. Also, since the embeddings are probability vectors, we use cosine distance between the extracted embeddings. Note that, we can use any other intermediate layer for embedding and Euclidean distance for measuring their separation. 

Since our objective is to generate diverse UAPs that can fool effectively, we give equal weight to both the components of the loss, i.e., we keep $\lambda =1$ in eq.~(\ref{eqn:loss}).
\subsection{UAPs and the success rates}
\label{subsec:uaps-success-rates}
Similar to~\cite{universal-cvpr-2017,mopuri-bmvc-2017,mopuri-cvpr-2018,mopuri-pami-2018} we measure the effectiveness of the crafted UAPs in terms of their ``success rate". It is the percentage of data samples $(x)$ for which the target CNN predicts a different label upon adding the UAP $(v)$. Note that we compute the success rates over the $50000$ validation images from ILSVRC dataset. Table~\ref{tab:fooling} reports the obtained success rates of the UAPs crafted by our generative model $G$ on various networks. Each row denotes the target model for which we train $G$ and the columns indicate the model we attack to fool. Thus, we report the transfer rates on the unseen models also, which is referred to as ``black-box attacking" (off-diagonal entries). Similarly, when the target CNN over which we learn $G$ matches with the model under attack, it is referred to as ``white-box attacking" (diagonal entries). Note that the right most column shows the mean success rates achieved by the individual generator networks $(G)$ obtained across all the $6$ CNN models. Proposed method can craft UAPs that have on an average $20.18\%$ higher mean success rate compared to the existing data-free method to craft UAPs (FFF~\cite{mopuri-bmvc-2017}). 

\begin{table*}[t]
\scriptsize
\centering
\caption{Success rates of the perturbations modelled by our generative network, compared against the data-free approach FFF~\cite{mopuri-bmvc-2017}. Rows indicate the target net for which perturbations are modelled and columns indicate the net under attack. Note that, in each row, entry where the target CNN matches with the network under attack represents white-box attack and the rest represent the black-box attacks. The mean fooling rate achieved by the Generator $(G)$ trained for each of the target CNNs is shown in the rightmost column.
}
\label{tab:fooling}
\begin{tabular}{|l|c|c|c|c|c|c|c|c|}
\cline{3-9}
\multicolumn{2}{l|}{}       & VGG-F          & CaffeNet       & GoogLeNet      & VGG-16         & VGG-19         & ResNet-152    & Mean FR \\ \cline{3-9} \hline 
\multirow{2}{*}{VGG-F}      & Ours & \textbf{92.37}  & \textbf{70.12}          & \textbf{58.51}& \textbf{47.01}          & \textbf{52.19} &  \textbf{43.22}  &  \textbf{60.56}      \\ \cline{2-9} 
                            & FFF &    81.59        &48.20            &38.56           &39.31           &39.19            &29.67           & 46.08 \\ \hline \hline
\multirow{2}{*}{CaffeNet}   & Ours & \textbf{74.68}          & \textbf{89.04} & \textbf{52.74}        & \textbf{50.39}          & \textbf{53.87}          & \textbf{44.63}  &    \textbf{60.89}    \\ \cline{2-9} 
                            & FFF &     56.18       &  80.92        &   39.38         &     37.22      &    37.62        &      26.45    &   46.29        \\ \hline \hline
\multirow{2}{*}{GoogLeNet}  & Ours & \textbf{57.90}         & \textbf{62.72}          & \textbf{75.28} & \textbf{59.12}          & \textbf{48.61}         & \textbf{47.81} &    \textbf{58.57}     \\ \cline{2-9} 
                        & FFF &      49.73      &   46.84         &     56.44       &   40.91         &     40.17       & 25.31      &  43.23        \\ \hline \hline
\multirow{2}{*}{VGG-16}     & Ours &  \textbf{58.27}        & \textbf{56.31}          & \textbf{60.74} & \textbf{71.59}         &  \textbf{65.64}       & \textbf{45.33}          & \textbf{59.64} \\ \cline{2-9} 
                        & FFF & 46.49           &   43.31         &     34.33       & 47.10           &     41.98       & 27.82    &     40.17 \\ \hline \hline
\multirow{2}{*}{VGG-19}     & Ours & \textbf{62.49}          & \textbf{59.62}         & \textbf{68.79}         & \textbf{69.45}         & \textbf{72.84} & \textbf{51.74}    &   \textbf{64.15} \\ \cline{2-9} 
                & FFF &   39.91       & 37.95            &      30.71      &    38.19       &       43.62     &     26.34      &    36.12     \\ \hline \hline
\multirow{2}{*}{ResNet-152} & Ours & \textbf{52.11}          & \textbf{57.16}         & \textbf{56.41}          & \textbf{47.21}          & \textbf{48.78}                & \textbf{60.72} & \textbf{53.73}\\ \cline{2-9} 
                            & FFF &      28.31      &     29.67       &     23.48       &      19.23      &      17.15     & 29.78  &    24.60       \\ \hline
\end{tabular}
\end{table*}
%

\begin{figure}[t]
\centering
\noindent\begin{minipage}{\textwidth}
  \centering
  \begin{minipage}{.19\textwidth}
  	\centering
    \includegraphics[width=\linewidth]{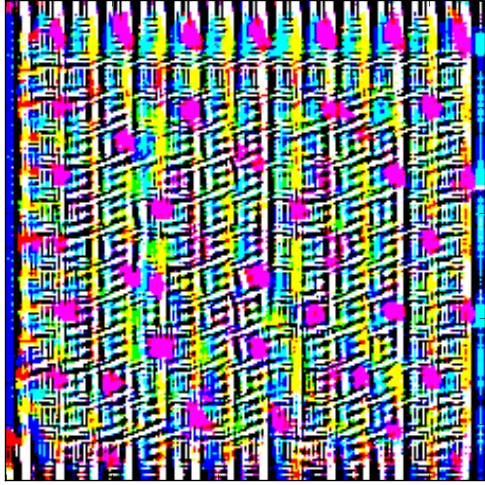}\
    CaffeNet
  \end{minipage}
   \begin{minipage}{.19\textwidth}
   	\centering
    \includegraphics[width=\linewidth]{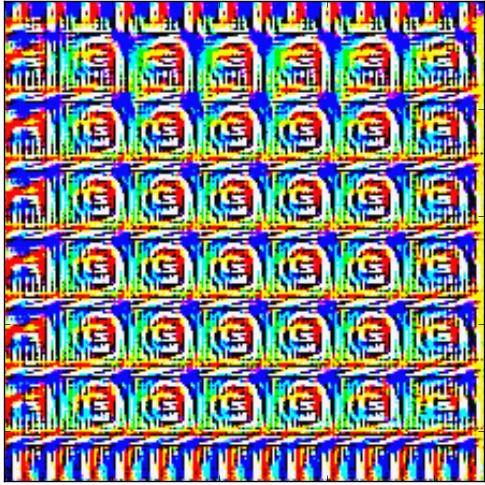}\
    VGG-F
  \end{minipage}
  \begin{minipage}{.19\textwidth}
  	\centering
    \includegraphics[width=\linewidth]{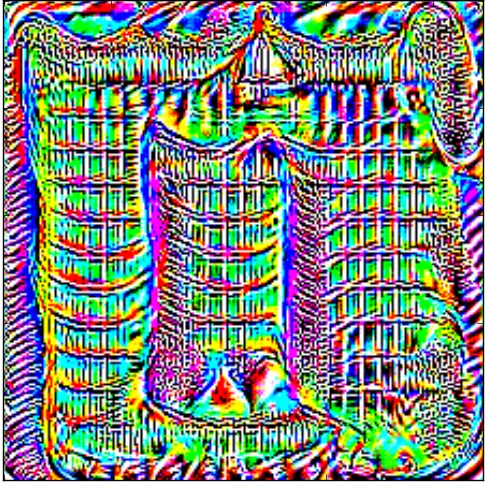}\
    GoogLeNet
  \end{minipage}
  \begin{minipage}{.19\textwidth}
  	\centering
    \includegraphics[width=\linewidth]{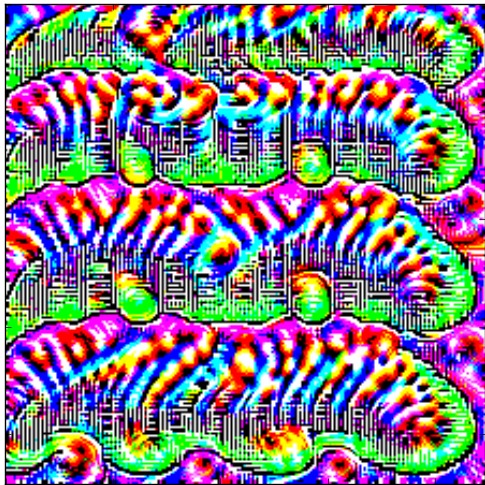}\
    VGG-19
  \end{minipage}
  \begin{minipage}{.19\textwidth}
  	\centering
    \includegraphics[width=\linewidth]{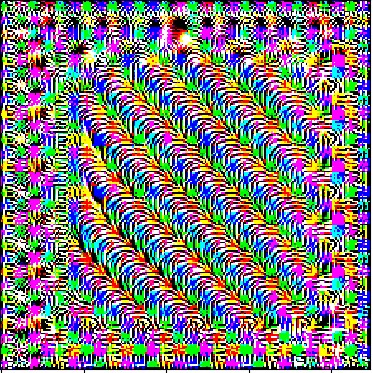}\
    ResNet-152
  \end{minipage}
\end{minipage}
\caption{Sample universal adversarial perturbations (UAP), learned by the proposed framework for different networks, the corresponding target CNN is mentioned below the UAP. Note that images shown are one sample for each of the target networks, and across different samplings the perturbations vary visually as shown in Fig.~\ref{fig:interpolation}.
}
\label{fig:sample_uaps}
\end{figure}
\begin{figure}[h]
\centering
\noindent\begin{minipage}{\textwidth}
  \centering
  \begin{minipage}{.19\textwidth}
  	\centering
    \includegraphics[width=\linewidth]{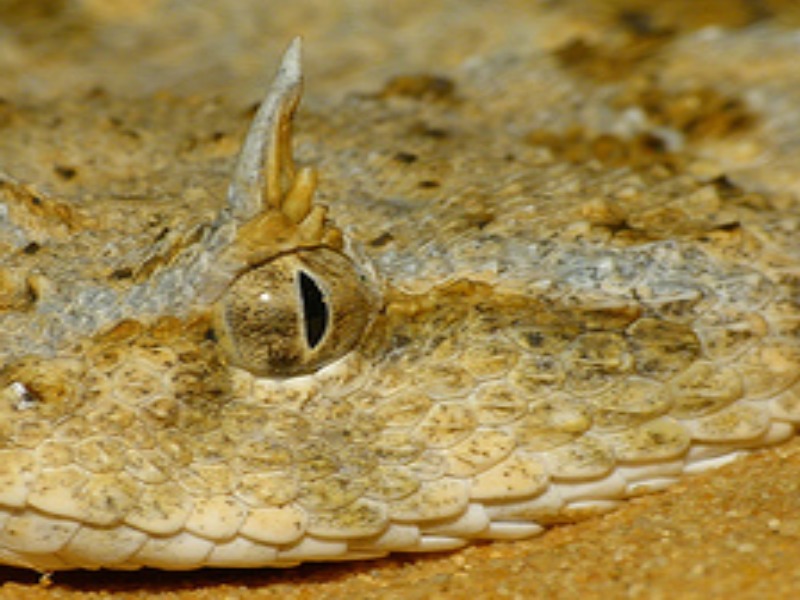}\
    Clean: Sand Viper
  \end{minipage}
   \begin{minipage}{.19\textwidth}
   	\centering
    \includegraphics[width=\linewidth]{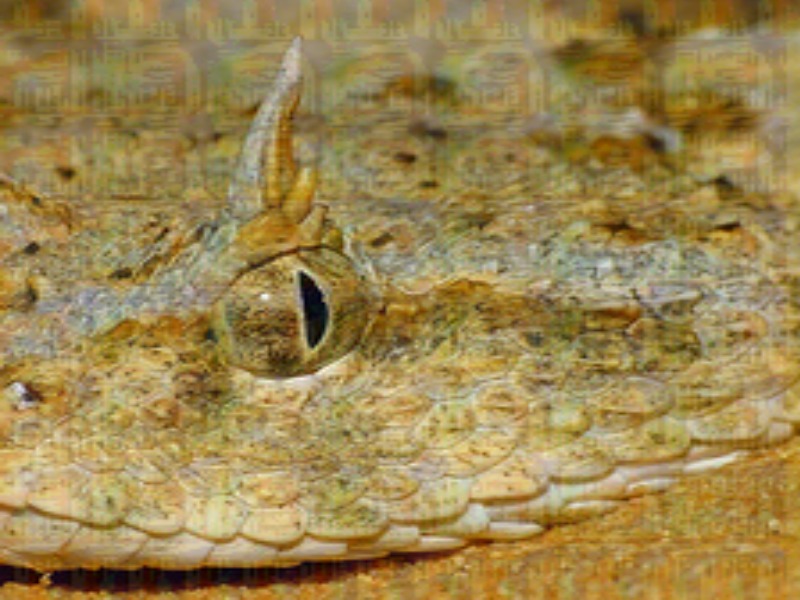}\
    VGG-F: Maypole
  \end{minipage}
  \begin{minipage}{.19\textwidth}
  	\centering
    \includegraphics[width=\linewidth]{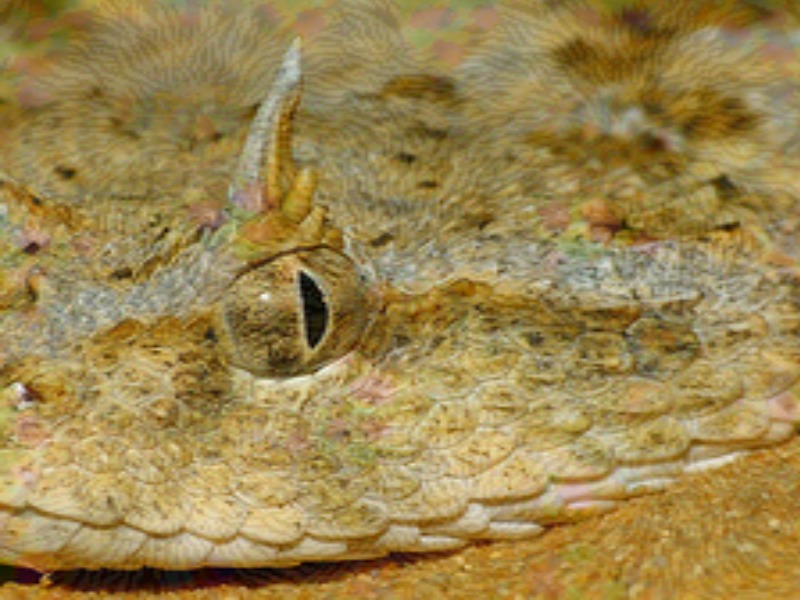}\
    CaffeNet: Afghan Hound
  \end{minipage}
  \begin{minipage}{.19\textwidth}
  	\centering
    \includegraphics[width=\linewidth]{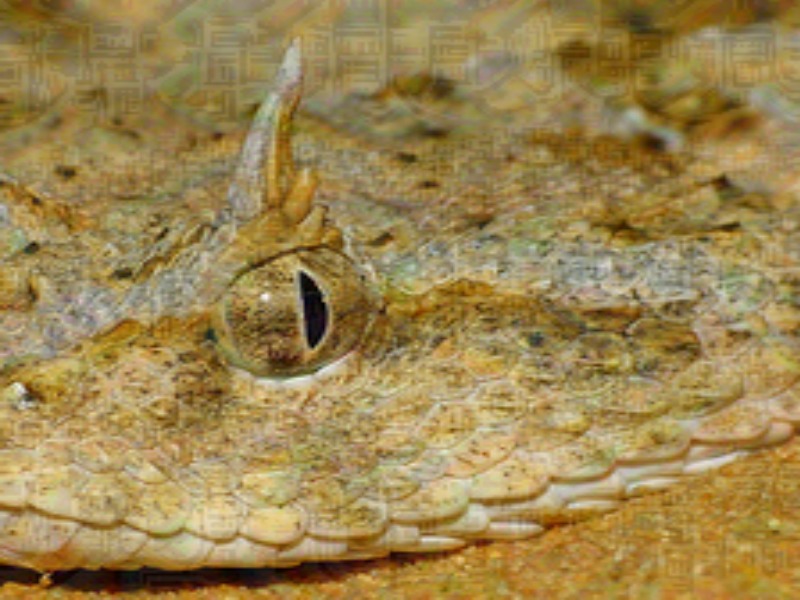}\
    VGG19: Egyptian Cat
  \end{minipage}
  \begin{minipage}{.19\textwidth}
  	\centering
    \includegraphics[width=\linewidth]{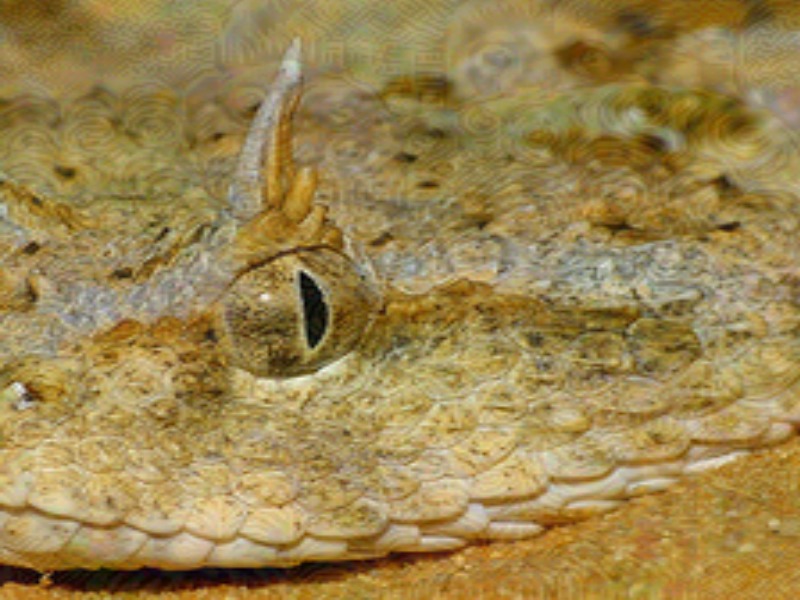}\
    ResNet152: Chiton
  \end{minipage}
\end{minipage}
\caption{Clean image (leftmost) of class ``Sand Viper", followed by adversarial images generated by adding UAPs crafted for various target CNNs. Note that the perturbations while remaining imperceptible are leading to different misclassifications.}
\label{fig:diversity}
\end{figure}

Figure~\ref{fig:sample_uaps} shows example UAPs learned by our approach for different target CNN models. Note that the pixel values in those perturbations lie in $[-10,10]$. Also the UAPs for different models look different. Figure~\ref{fig:diversity} shows a clean and corresponding perturbed samples after adding UAPs learned for different target CNNs. Note that each of the target CNNs misclassify them differently.

For the sake of completeness, we compare our approach with the data-driven counterpart also. Table~\ref{tab:comparison} presents the white-box success rates for both data-free and data-driven methods to craft UAPs. We also show the fooling ability of random noise sampled in $[-10,10]$ as a baseline. Note that the success rates obtained by random noise is very less compared to the learned UAPs. Thus the adversarial perturbations are highly structured and very effective compared to the performance of random noise as perturbation.

On the other hand, the proposed method of acquiring class impressions from the target model's memory increases the mean success rate by an absolute $20\%$ from that of current state-of-the-art data-free approach (FFF~\cite{mopuri-bmvc-2017}). Also, note that our approach performs close to the data-driven approach UAP~\cite{universal-cvpr-2017} with a gap of $8\%$. These observations suggest that the class impressions are effective to serve the purpose of the actual data samples in the context of learning to craft the UAPs. 
\begin{table}[]
\centering
\caption{Effectiveness of the proposed approach to handle the data absence. We compare the success rates against the data-driven approach UAP~\cite{universal-cvpr-2017}, data-free approach FFF~\cite{mopuri-bmvc-2017} and random noise baseline.
}
\label{tab:comparison}
\begin{tabular}{|l|c|c|c|c|c|c|c|}
\hline
 & VGG-F & CaffeNet & GoogLeNet & VGG-16 & VGG-19 & ResNet-152 & Mean\\ \hline
Baseline          &  12.62     &    12.9      &     10.29      &    8.62    &  8.40      &      8.99 &   10.30   \\ \hline
FFF (w/o Data)               &  81.59      &   80.92       &       56.44    &      47.10  &    43.62    &      29.78  &  56.58  \\ \hline
Ours(w/o Data)              &  92.37      &       89.04   &  75.28         &      71.59  &    69.45    &      60.72   &  76.41 \\ \hline
UAP (w Data)              &  93.8      &    93.1      &     78.5      &     77.8   &    80.8    &  84.0 & 84.67   \\ \hline
\end{tabular}
\end{table}

\subsection{Comparison with data dependent approaches.}
\label{sup_sec:transfer-rates}
Table~\ref{sup_tab:transferability} presents the transfer rates achieved by the image-agnostic perturbations crafted by the proposed approach. Each row denotes the target model on which the generative model $(G)$ is learned and columns denotes the models under attack. Hence, diagonal entries denote the white-box adversarial attacks and the off diagonal entries denote the black-box attacks. Note that the main draft presents only the white-box success rates, for completeness we present both here. Also note that, in spite of being a data-free approach the mean SR (extreme right column) obtained by our method is very close to that achieved by the state-of-the-art data-driven approach to craft UAPs.

\begin{table}
\scriptsize
\centering
\caption{Success rates (SR) for the perturbations crafted by the proposed approach compared against the state-of-the-art data driven approach for crafting the UAPs.
}

\label{sup_tab:transferability}
\begin{tabular}{|l|c|c|c|c|c|c|c|c|}
\cline{3-9}
\multicolumn{2}{l|}{}                       & VGG-F          & CaffeNet       & GoogLeNet      & VGG-16         & VGG-19         & ResNet-152    & Mean SR \\ \cline{3-9} \hline 
\multirow{2}{*}{VGG-F}      & Ours &        92.37&              70.12        &  58.51 &         47.01&              52.19 &         43.22  &        60.56      \\ \cline{2-9} 
                            & UAP &         93.7 &              71.8        &   48.4        &   42.1            &   42.1        &   47.4        &   57.58 \\ \hline \hline
\multirow{2}{*}{CaffeNet}   & Ours &        74.68          &    89.04 &         52.74        &  50.39          &    53.87         & 44.63  &        60.89    \\ \cline{2-9} 
                            & UAP &         74.0        &       93.3         &  47.7          & 39.9       &        39.9         &  48.0     &      56.71        \\ \hline \hline
\multirow{2}{*}{GoogLeNet}  & Ours &        57.90         &     62.72          &75.28 &         59.12          &    48.61         & 47.81 &         58.57     \\ \cline{2-9} 
                            & UAP &         46.2       &        43.8          & 78.9        &   39.2          &     39.8        &   45.5       &    48.9         \\ \hline \hline
\multirow{2}{*}{VGG-16}     & Ours &        58.27        &      56.31          &60.74 &         71.59         &     65.64       &   45.33        &  59.64 \\ \cline{2-9} 
                            & UAP &         63.4            &   55.8          & 56.5        &   78.3            &   73.1        &   63.4     &      65.08 \\ \hline \hline
\multirow{2}{*}{VGG-19}     & Ours &        62.49          &    59.62         & 68.79         & 69.45         &     72.84 &         51.74    &      64.15 \\ \cline{2-9} 
                            & UAP &         64.0        &       57.2    &       53.6       &    73.5        &       77.8      &     58.0       &    64.01     \\ \hline \hline
\multirow{2}{*}{ResNet-152} & Ours &        52.11          &    57.16         & 56.41       &   47.21          &    48.78   &       60.72 &         53.73\\ \cline{2-9} 
                        & UAP &         46.3       &        46.3        &   50.5        &   47.0       &        45.5      &     84.0   &        53.27       \\ \hline
\end{tabular}
\end{table}


\subsection{Diversity}
\label{subsec:diversity}
The objective of having the diversity component $(L_d)$ in the loss is to avoid learning a single UAP and to learn a generative model that can generate diverse set of UAPs for a given target CNN. We examine the distribution of predicted labels after adding the generated UAPs. This can reveal if there is a set of sink labels that attract most of the predictions. We have considered the $G$ learned to fool VGG-F model and $50000$ samples of ILSVRC validation set. We randomly select $10$ UAPs generated by the $G$ and compute the mean histogram of predicted labels. After sorting the histogram, most of the predicted labels $(95\%)$ for proposed approach spread over $212$ labels out of the total $1000$ target labels. Whereas the same number for UAP~\cite{universal-cvpr-2017} is $173$. The observed $22.5\%$ higher diversity is attributed to our diversity component $(L_d)$. 

\subsection{Simultaneous Targets}
\label{subsec:simultaneous-targets}
The ability of the adversarial perturbations to generalize across multiple models is observed with both image-specific (\cite{intriguing-iclr-2014,explainingharnessing-iclr-2015}) and agnostic perturbations (\cite{universal-cvpr-2017,mopuri-bmvc-2017}). It is an important issue to be investigated since it makes simple black-box attacks possible via transferring the perturbations to unknown models. In this subsection we investigate to learn a single $G$ that can can craft UAPs to simultaneously fool multiple target CNNs.

We replace the single target CNN with an ensemble of three models: CaffeNet, VGG-16 and ResNet-152 and learn $G_E$ using the fooling and diversity losses. Note that, since the class impressions vary from model to model, for this experiment we generate class impressions from multiple CNNs. Particularly, we simultaneously maximize the pre-softmax activation (eq.(~\ref{eqn:class-impressions})) of the desired class across individual target CNNs via optimizing their mean. We then investigate the generalizability of the generated perturbations. Table~\ref{tab:multi-targets} presents the mean black-box success rate (MBBSR) for the UAPs generated by $G_E$ on the remaining $3$ models. For comparison, we present the MBBSR of the generators learned on the individual models. Because of the ensemble of the target CNNs $G_E$ learns to craft more general UAPs and therefore achieves higher success rates than the individual generators.
\begin{table}[t]
\centering
\caption{Generalizability of the UAPs crafted by the ensemble generator $G_E$ learned on three target CNNs: CaffeNet, VGG-16 and ResNet-152. Note that because of the ensemble of the target CNNs, $G_E$ learns to craft perturbations that have higher mean black-box success rates (MBBSR) compared to that of the individual generators. }
\label{tab:multi-targets}
\begin{tabular}{|l|l|l|l|l|}
\hline
      & $G_C$ & $G_{V16}$ & $G_{R152}$ & $G_E$ \\ \hline
MBBSR &   60.34   &     61.46       &   52.43          &    \textbf{68.52}      \\ \hline
\end{tabular}
\end{table}

\subsection{Interpolating in the latent space}
\label{subsec:interpolation}
Our generator network $(G)$ is similar to that in a typical GAN~\cite{gan-nips-2014,dcgan-arxiv-2015}. It maps the latent space to the space of UAPs for the given target classifier(s). In case of GANs, interpolating in the latent space can reveal signs of memorization. While traversing the latent space, smooth semantic change in the generations means the model has learned relevant representations. In our case, since we generate UAPs, we investigate if the interpolation has smooth visual changes and the intermediate UAPs can also fool the target CNN coherently.

\begin{figure}[h]
\centering
\noindent\begin{minipage}{\textwidth}
  \centering
  \begin{minipage}{.18\textwidth}
  	\centering
    \includegraphics[width=\linewidth]{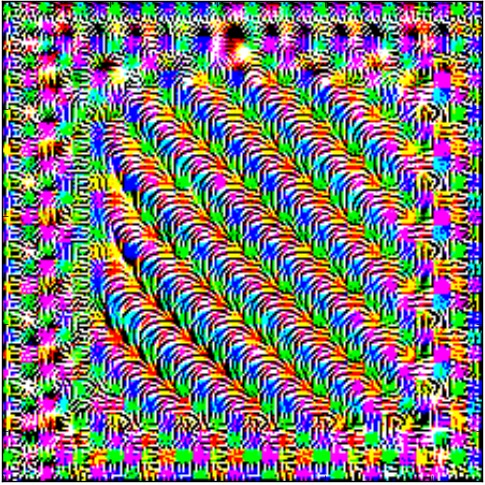}\
    $0.0*z_1+1.0*z_2:60.58$
  \end{minipage}
   \begin{minipage}{.18\textwidth}
   	\centering
    \includegraphics[width=\linewidth]{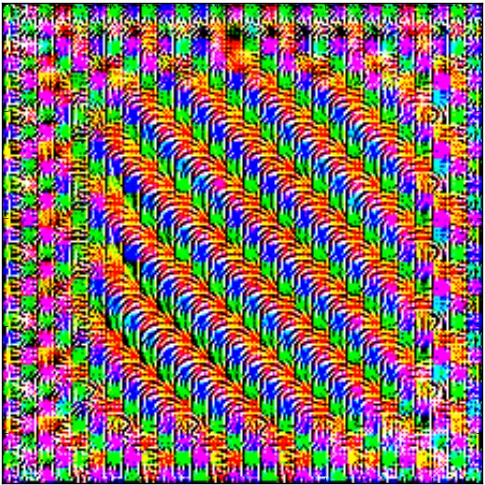}\
    $0.25*z_1+0.75*z_2:59.16$
  \end{minipage}
  \begin{minipage}{.18\textwidth}
  	\centering
    \includegraphics[width=\linewidth]{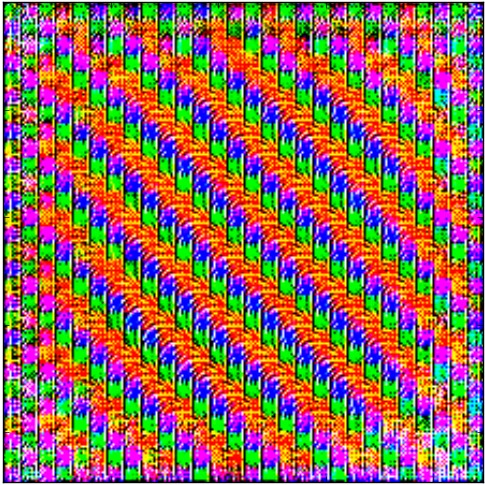}\
    $0.5*z_1+0.5*z_2:60.25$
  \end{minipage}
   \begin{minipage}{.18\textwidth}
   	\centering
    \includegraphics[width=\linewidth]{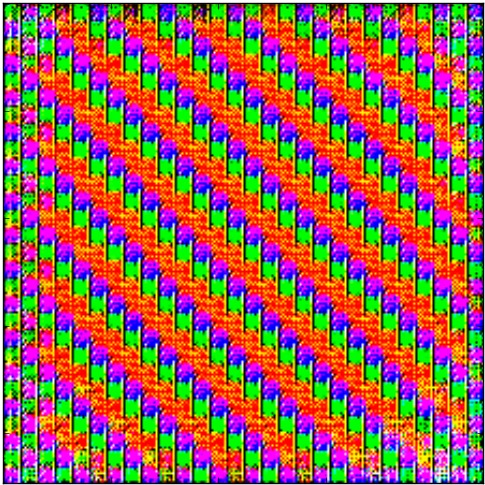}\
    $0.75*z_1+0.25*z_2:59.87$
  \end{minipage}
  \begin{minipage}{.18\textwidth}
  	\centering
    \includegraphics[width=\linewidth]{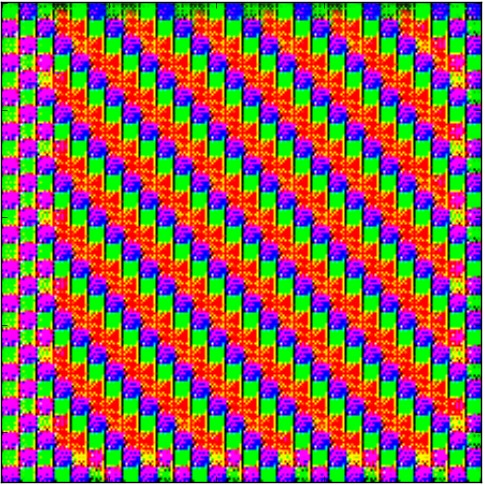}\
     $1.0*z_1+0.0*z_2:60.09$
  \end{minipage}
\end{minipage}
\caption{Interpolation between a pair of points in $Z$ space shows that the mapping learned by our generator has smooth transitions. The figure shows the perturbations corresponding to $5$ points on the line joining a pair of points $(z_1 \text{ and } z_2)$ in the latent space. Note that these perturbations are learned to fool the ResNet-152~\cite{resnet-2015} architecture. Below each perturbation, the corresponding success rate obtained over $50000$ images from ILSVRC $2014$ validation images is mentioned. This shows the fooling capability of these intermediate perturbations is also high and remains same at different locations.}
\label{fig:interpolation}
\end{figure}

Figure~\ref{fig:interpolation} shows the results of interpolating in the latent space for ResNet-152 as the target CNN. We sample a pair of points $(z_1\text{ and }z_2)$ in the latent space and consider $5$ intermediate points on the line joining them. We generate the UAPs corresponding to all these points by passing them through the learned generator architecture $G$. Figure~\ref{fig:interpolation} shows the generated UAPs and the corresponding success rates in fooling the target CNN. Note that the UAPs change visually smoothly between any pair of points and the success rate remains unchanged. This ensures that the representations learned are relevant and interesting.

\subsection{Adversarial Training}
\label{subsec:at}
We have performed adversarial training of target CNN with $50\%$ mixture of clean and adversarial samples crafted using the learned generator (G). After $2$ epochs, success rate of the G has dropped from $75.28$ to $62.51$. Note that the improvement is minor and the target CNN is still vulnerable. We then repeated the generator training for the finetuned network, resulting generator fools the finetuned network with an increased success rate of $68.72$. After repeating this for multiple iterations, we observe that adversarial training does not make the target CNN significantly robust.\\

\section{Discussion and Conclusions}
\label{sec:conclusions}
In this paper we have presented a novel approach to mitigate the absence of data for crafting Universal Adversarial Perturbations (UAP). Class impressions are representative images that are easy to obtain via simple optimization from the target model. Using class impressions, our method drastically reduces the performance gap between the data-driven and data-free approaches to craft the UAPs. Success rates closer to that of data-driven UAPs demonstrate the effectiveness of class impressions in the context of crafting UAPs.

Another way to look at this observation is that it would be possible to extract useful information about the training data from the model parameters in a task specific manner. In this paper, we have extracted the class impressions as proxy data samples to train a generative model that can craft UAPs for the given target CNN classifier. It would be interesting to explore such feasibility for other applications as well. Particularly, we would like to investigate if the existing adversarial setup of the GANs might get benefited with any additional information extracted from the discriminator network and generate more natural looking synthetic data. 

The generative model presented in our approach is an efficient way to craft UAPs. Unlike the existing methods that perform complex optimizations, our approach constructs UAPs through a simple feed forward operation. Significant success rates, surprising cross model generalizability even in the absence of data reveal severe susceptibilities of the current deep learning models. 
\bibliographystyle{splncs04}
\bibliography{mybibliography}

\begin{thebibliography}{10}
\providecommand{\url}[1]{\texttt{#1}}
\providecommand{\urlprefix}{URL }
\providecommand{\doi}[1]{https://doi.org/#1}

\bibitem{tensorflow2015-whitepaper-short}
Abadi~et al., M.: {TensorFlow}: Large-scale machine learning on heterogeneous
  systems (2015), \url{http://tensorflow.org/}, software available from
  tensorflow.org

\bibitem{atn-aaai-2018}
Baluja, S., Fischer, I.: Learning to attack: Adversarial transformation
  networks. In: Proceedings of AAAI (2018)

\bibitem{evasion-mlkd-2013}
Biggio, B., Corona, I., Maiorca, D., Nelson, B., {\v{S}}rndi{\'c}, N., Laskov,
  P., Giacinto, G., Roli, F.: Evasion attacks against machine learning at test
  time. In: Joint European Conference on Machine Learning and Knowledge
  Discovery in Databases. pp. 387--402 (2013)

\bibitem{prsystemsunderattack-pari-2014}
Biggio, B., Fumera, G., Roli, F.: Pattern recognition systems under attack:
  Design issues and research challenges. International Journal of Pattern
  Recognition and Artificial Intelligence  \textbf{28}(07) (2014)

\bibitem{vggf-bmvc-2014}
Chatfield, K., Simonyan, K., Vedaldi, A., Zisserman, A.: Return of the devil in
  the details: Delving deep into convolutional nets. In: Proceedings of the
  British Machine Vision Conference {(BMVC)} (2014)

\bibitem{gan-nips-2014}
Goodfellow, I.J., Pouget-Abadie, J., Mirza, M., Xu, B., Warde-Farley, D.,
  Ozair, S., Courville, A., Bengio, Y.: Generative adversarial nets. In:
  Advances in {N}eural {I}nformation {P}rocessing {S}ystems, {(NIPS)} (2014)

\bibitem{explainingharnessing-iclr-2015}
Goodfellow, I.J., Shlens, J., Szegedy, C.: Explaining and harnessing
  adversarial examples. In: International Conference on Learning
  Representations (ICLR) (2015)

\bibitem{resnet-2015}
He, K., Zhang, X., Ren, S., Sun, J.: Deep residual learning for image
  recognition. arXiv preprint arXiv:1512.03385  (2015)

\bibitem{adversarialml-acmmm-2011}
Huang, L., Joseph, A.D., Nelson, B., Rubinstein, B.I., Tygar, J.D.: Adversarial
  machine learning. In: Proceedings of the 4th ACM Workshop on Security and
  Artificial Intelligence. AISec '11 (2011)

\bibitem{kingma2014adam}
Kingma, D., Ba, J.: Adam: A method for stochastic optimization. arXiv preprint
  arXiv:1412.6980  (2014)

\bibitem{physicalworld-iclr-2017}
Kurakin, A., Goodfellow, I., Bengio, S.: Adversarial examples in the physical
  world. In: International Conference on Learning Representations (ICLR) (2017)

\bibitem{delving-iclr-2017}
Liu, Y., Chen, X., Liu, C., Song, D.: Delving into transferable adversarial
  examples and black-box attacks. In: International Conference on Learning
  Representations (ICLR) (2017)

\bibitem{universal-cvpr-2017}
Moosavi{-}Dezfooli, S., Fawzi, A., Fawzi, O., Frossard, P.: Universal
  adversarial perturbations. In: IEEE Conference on Computer Vision and Pattern
  Recognition (CVPR) (2017)

\bibitem{deepfool-cvpr-2016}
Moosavi{-}Dezfooli, S., Fawzi, A., Frossard, P.: Deepfool: {A} simple and
  accurate method to fool deep neural networks. In: IEEE conference on Computer
  Vision and Pattern Recognition {(CVPR)} (2016)

\bibitem{mopuri-pami-2018}
Mopuri, K.R., Ganeshan, A., Babu, R.V.: Generalizable data-free objective for
  crafting universal adversarial perturbations. IEEE Transactions on Pattern
  Analysis and Machine Intelligence  (2018)

\bibitem{mopuri-cnnf-2017}
Mopuri, K.R., Garg, U., Babu, R.V.: {CNN} fixations: An unraveling approach to
  visualize the discriminative image regions. arXiv preprint arXiv:1708.06670
  (2017)

\bibitem{mopuri-bmvc-2017}
Mopuri, K.R., Garg, U., Babu, R.V.: Fast feature fool: A data independent
  approach to universal adversarial perturbations. In: Proceedings of the
  British Machine Vision Conference ({BMVC}) (2017)

\bibitem{mopuri-cvpr-2018}
Mopuri, K.R., Ojha, U., Garg, U., Babu, R.V.: {NAG}: Network for adversary
  generation. In: Proceedings of the IEEE conference on Computer Vision and
  Pattern Recognition ({CVPR}) (2018)

\bibitem{deep-dream-2015}
Mordvintsev, A., Tyka, M., Olah, C.: Google deep dream  (2015),
  \url{https://research.googleblog.com/2015/06/inceptionism-going-deeper-into-neural.html}

\bibitem{olah2017feature}
Olah, C., Mordvintsev, A., Schubert, L.: Feature visualization. Distill
  (2017), \url{https://distill.pub/2017/feature-visualization}

\bibitem{practicalbb-asiaccs-2017}
Papernot, N., McDaniel, P.D., Goodfellow, I.J., Jha, S., Celik, Z.B., Swami,
  A.: Practical black-box attacks against deep learning systems using
  adversarial examples. In: Asia Conference on Computer and Communications
  Security (ASIACCS) (2017)

\bibitem{dcgan-arxiv-2015}
Radford, A., Metz, L., Chintala, S.: Unsupervised representation learning with
  deep convolutional generative adversarial networks. arXiv preprint
  arXiv:1511.06434  (2015)

\bibitem{imagenet-ijcv-2015}
Russakovsky, O., Deng, J., Su, H., Krause, J., Satheesh, S., Ma, S., Huang, Z.,
  Karpathy, A., Khosla, A., Bernstein, M., Berg, A.C., Fei-Fei, L.: {ImageNet
  Large Scale Visual Recognition Challenge}. International Journal of Computer
  Vision (IJCV)  \textbf{115}(3),  211--252 (2015)

\bibitem{NIPS2016_6125}
Salimans, T., Goodfellow, I.J., Zaremba, W., Cheung, V., Radford, A., Chen, X.:
  Improved techniques for training gans. In: Advances in Neural Information
  Processing Systems {(NIPS)} (2016)

\bibitem{gradcam-iccv-2017}
Selvaraju, R.R., Cogswell, M., Das, A., Vedantam, R., Parikh, D., Batra, D.:
  Grad-cam: Visual explanations from deep networks via gradient-based
  localization. In: The IEEE International Conference on Computer Vision (ICCV)
  (2017)

\bibitem{backprop-iclrw-2014}
Simonyan, K., Vedaldi, A., Zisserman, A.: Deep inside convolutional networks:
  Visualising image classification models and saliency maps. In: International
  Conference on Learning Representations {ICLR} Workshops (2014)

\bibitem{guidedbackprop-iclrw-2015}
Springenberg, J., Dosovitskiy, A., Brox, T., Riedmiller, M.: Striving for
  simplicity: The all convolutional net. In: International Conference on
  Learning Representations (ICLR) (workshop track) (2015)

\bibitem{intriguing-iclr-2014}
Szegedy, C., Zaremba, W., Sutskever, I., Bruna, J., Erhan, D., Goodfellow,
  I.J., Fergus, R.: Intriguing properties of neural networks. In: International
  Conference on Learning Representations (ICLR) (2013)

\bibitem{deconv-eccv-2014}
Zeiler, M.D., Fergus, R.: Visualizing and understanding convolutional networks.
  In: European Conference on Computer Vision ({ECCV}). pp. 818--833 (2014)

\bibitem{exbp-eccv-2016}
Zhang, J., Lin, Z., Brandt, J., Shen, X., Sclaroff, S.: Top-down neural
  attention by excitation backprop. In: European Conference on Computer
  Vision(ECCV) (2016)

\bibitem{cam-cvpr-2016}
Zhou, B., Khosla, A., Lapedriza, A., Oliva, A., Torralba, A.: Learning deep
  features for discriminative localization. In: Proceedings of Computer Vision
  and Pattern Recognition ({CVPR}) (2016)

\end{thebibliography}
\end{document}